\documentclass{article}

     \usepackage[preprint]{neurips_2019}

\usepackage[utf8]{inputenc} 
\usepackage[T1]{fontenc}    
\usepackage{hyperref}       
\usepackage{url}            
\usepackage{booktabs}       
\usepackage{amsfonts}       
\usepackage{nicefrac}       
\usepackage{microtype}      
\usepackage{multicol}
\usepackage{amsmath}
\usepackage{graphicx}
\usepackage{capt-of}
\usepackage{stmaryrd}
\usepackage{placeins}
\usepackage{float}
\title{Relational Graph Representation Learning for Open-Domain Question Answering}

\begin{document}

\author{Salvatore Vivona\\
  Autodesk AI Lab\\
  Toronto, Canada\\
  \texttt{vivonasg@cs.toronto.edu}
  \And
  Kaveh Hassani\\
  Autodesk AI Lab\\
  Toronto, Canada\\
  \texttt{kaveh.hassani@autodesk.com}
}

\maketitle
\begin{abstract}
We introduce a relational graph neural network with bi-directional attention mechanism and hierarchical representation 
learning for open-domain question answering task. Our model can learn contextual representation by jointly learning and 
updating the query, knowledge graph, and document representations. The experiments suggest that our model achieves 
state-of-the-art on the WebQuestionsSP benchmark.  
\end{abstract}
\section{Introduction}
Fusing structured knowledge from knowledge graphs into deep models using Graph Neural Networks (GNN) \cite{wu_2019_arxiv, zhou_2018_arxiv, zhang_2018_arxiv} is shown to improve their performance on tasks such as visual question answering
\cite{narasimhan_2018_neurips}, object detection \cite{marino_2017_cvpr}, natural language inference \cite{chen_2018_acl}, neural 
machine translation\cite{moussallem_2019_arxiv}, and open-domain question answering \cite{sun_2018_acl}. This particularly 
helps question answering neural models such as Memory Networks \cite{weston_2015_iclr} and Key-Value Memory Networks
\cite{miller_2016_emnlp} by providing them with well-structured knowledge on specific and open domains \cite{zhang_2018_aaai}. 
Most models, however, answer questions using a single information source, usually either a text corpus, or a single knowledge 
graph. Text corpora have high coverage but extracting information from them is challenging whereas knowledge graphs are 
incomplete but are easier to extract answers from \cite{sun_2018_acl}.

In this paper, we propose a relational GNN for open-domain question answering that learns contextual knowledge graph embeddings 
by jointly updating the embeddings from a knowledge graph and a set of linked documents. More specifically, our contributions are as 
follows: (1) we use documents as contextualized relations to augment the knowledge graph and co-train the knowledge graph and 
document representations,  (2) we introduce a bi-directional graph attention mechanism for knowledge graphs, (3) we propose a simple 
graph coarsening method to fuse node and cluster representations, and (4) we show that our model achieves state-of-the-art results on the
WebQuestionsSP benchmark.

\section{Related Works}
Graph representation learning on knowledge graphs allows projecting high-level factual information into embedding spaces which 
can be fused with other learned structural representations to improve down-stream tasks. Pioneer works such as Trans-E
\cite{bordes_2013_neurips}, Complex-E \cite{trouillon_2016_icml}, Hole-E\cite{nickel_2016_aaai}, and DistMul \cite{yang_2015_iclr} 
use unsupervised and mostly linear models to learn such pre-trained representations. A few recent works, on the other hand, use 
GNNs to compute the knowledge graph representation \cite{xie_2016_aaai,wang_2019_kdd,zhang_2019_neurips}. These high-level 
knowledge graph representations are particularly important for question answering task \cite{weston_2015_iclr,miller_2016_emnlp,
sun_2018_acl}. We use pre-trained representations to initialize the model and then update them using a relational and bi-directional
GNN model.

Only a few works fuse knowledge graphs with text corpora to answer questions. In \cite{das_2017_acl} early fusion of knowledge 
graph facts and text is performed using Key-Value Memory Networks (KVMN). This model, however, ignores relational structure between 
the text and knowledge graph. Our model links the knowledge graph and documents through document-contextualized edges and 
also links entities with their positions in the corpus. This linking is used in GRAFT-Net as well which also performs question answering 
through fusing learned knowledge graph and linked document representations \cite{sun_2018_acl}.  Unlike GRAFT-Net, our model uses 
variants of differential pooling \cite{ying_2018_neurips} and bi-directional graph attention \cite{velickovic_2018_iclr} for more 
powerful message passing. Our model also introduces trainable document-contextualized relation embeddings instead of exclusively 
relying on fixed relation representations.

\section{Method}
\label{sec:methodology}
Assume a knowledge graph $G=\left(V, E, R\right)$ where $V$, $R$, and $E$ are sets of entities, relations, and edges, respectively. Each edge
$e_i=(s,r,o)\in E$ denotes an object $o$ interacting with subject $s$ through a relationship $r$. Given a textual query $q \in {(w_1...w_{|q|})}$  
and a set of relevant documents $D$ linked to $G$ by an imperfect linker, the task is to predict the answer nodes: $v_{a_q}\in V$. A query can 
have zero or multiple answers and hence the task is reduced to binary node classification on $G$ (i.e., binary cross entropy loss). 
Following \cite{sun_2018_acl}, we first extract a subgraph $G_q \subset G$ which contains $v_{a_q}$ with high probability. This is done by linking 
the query entities to $G$ and expanding their neighborhood using Personalized PageRank (PPR) method. We then initialize $G_q$, $q$, $D$, 
and $R$ in embedding space as follows. Each entity $v\in V$ is initialized using pre-trained TransE embeddings \cite{bordes_2013_neurips}: 
$h_v^{(0)}=\text{TransE}(v) 
\in {\mathbb{R}^{d_{kb}}}$. Documents are encoded as $H_D^{(l)} \in \mathbb{R}^{|D|\times n_d\times d_w}$ where $n_d$ is the maximum number of tokens in documents, and $H_{D_{k,j}}^{(l)}\in \mathbb{R}^{d_w}$ corresponds to the $j$th token embedding of dimension $d_w$ in the $k$th document. A bidirectional LSTM that takes in pre-trained GloVe embeddings 
\cite{penn_2016_emnlp} of tokens and computes a global sequence embedding is shared between query and documents to initialize 
their embeddings: $h^{(0)}_q=\text{BiLSTM}\left(w_1...w_{|q|}\right)$ and $H_d^{(0)}=\text{BiLSTM}\left(w_1...w_{|d|}\right)$. Each relation $r\in R$ is initialized as:
$h_r^{(0)}=\text{ReLU}\left(\textbf{W}_r^{(0)}\left[\text{TransE}(r)\parallel \mu_{glove}(w_1...w_{|r|}) \right]\right)$ where $\textbf{W}_r^{(0)}\in \mathbb{R}^{(d_{kb}+d_w)\times d_r}$ is a trainable parameter, $\parallel$ is the concatenation operator, and $\mu_{glove}\in \mathbb{R}^{d_w}$ 
is the mean pool over GloVe embeddings of the relation tokens.

\subsection{Documents as Relations}
\begin{figure}
    \centering
  \includegraphics[width=100mm,scale=0.5]{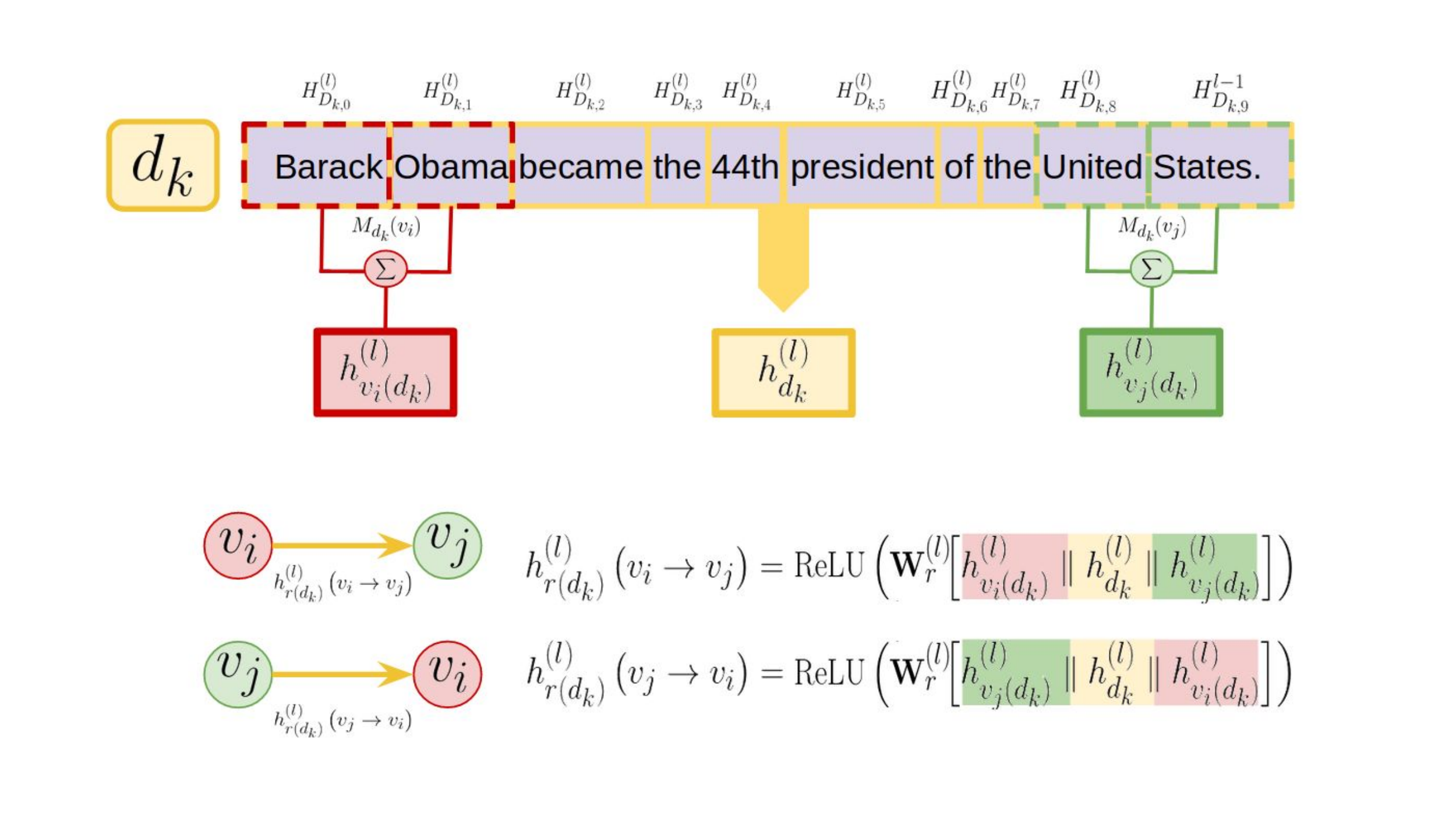}
  \caption{An example of forming relations from documents.}
  \label{fig:DOCEDGE}
\end{figure}

We use documents to explicitly augment the knowledge graph by forming additional relations between entities. Assuming two unique 
entities $v_i$ and $v_j$ co-occurring within document $d_k$, we compute the document relation embedding for both plausible directions between 
$v_i$ and $v_j$. This lets the model to learn to attend to the direction that optimizes the objective. 
These embeddings are computed as follows:
\begin{equation} 
h^{(l)}_{r(d_k)}\left(v_i\rightarrow v_j\right) = \text{ReLU}\left(\textbf{W}_{dr}^{(l)}\left[h_{v_i(d_k)}^{(l)} \parallel  h^{(l)}_{d_k} \parallel
h_{v_j(d_k)}^{(l)}\right]\right)
\tag{1a}\label{eq:3a}
\end{equation}
\begin{equation} 
h^{(l)}_{r(d_k)}(v_j\rightarrow v_i) = \text{ReLU}\left(\textbf{W}_{dr}^{(l)}\left[h_{v_j(d_k)}^{(l)} \parallel  h^{(l)}_{d_k} \parallel
h_{v_i(d_k)}^{(l)}\right]\right)
\tag{1b}\label{eq:3b}
\end{equation}
where $\textbf{W}_{dr}^{(l)}\in \mathbb{R}^{3d_w\times d_r}$ is a trainable parameter, $h_{d_k}^{(l)}$ is the learned textual embedding of the document and $h_{v(d_k)}^{(l)}=\sum_{p\in M_{d_k}(v)}{H_{d_k,p}^{(l)}}$ denotes the textual representation of the entity 
and is computed by summing up the textual embeddings of its tokens within the document (i.e., $M_{d_k}(v)$ returns the positions of 
entity $v$ in document $d_k$). An example of this process is illustrated in Figure \ref{fig:DOCEDGE}.
 
\subsection{Bi-Directional Graph Attention}
\label{BDRU}
To update the node embedding $h^{(l)}_{v}$ we aggregate the embeddings of its connecting edges. An edge embedding is computed by 
aggregating its relation embedding $h_{r}^{(l)}$ with the neighbor node embedding $h_{v}^{(l)}$ connecting through that edge. Let $e_{out}=(v,r,v_i)\in E$ 
and $e_{in}=(v_j,r,v) \in E$ represent two facts in which the node $v$ is either the subject or the object. If the node is an object, we will 
refer the edge to point inwards, and outwards if it is a subject. The edge embedding are computed as follows:
\begin{equation}
	h^{(l)}_{e_{out}}=\text{ReLU}\left(\textbf{W}_r^{(l)}h^{(l)}_{r} + \textbf{W}_v^{(l)}h^{(l)}_{v_i}\right) ,\quad
    h^{(l)}_{e_{in}}=\text{ReLU}\left(\textbf{W}_r^{(l)}(-h^{(l)}_{r}) +  \textbf{W}_v^{(l)}h^{(l)}_{v_j}\right)
    \tag{2}\label{eq:4}
\end{equation}
where $h_{e_{in}}^{(l)}$ and $h_{e_{out}}^{(l)}$ denote the embeddings of the inward and outward edges connecting to node $v$, and $\textbf{W}_r^{(l)}$ and $\textbf{W}_v^{(l)}$ are trainable 
parameters. To distinguish inward edges from outward edges, we negate $h_{r}$. This is distinct from previous approaches 
which only process incoming nodes \cite{sun_2018_acl}.

\begin{figure}
  \includegraphics[width=\linewidth]{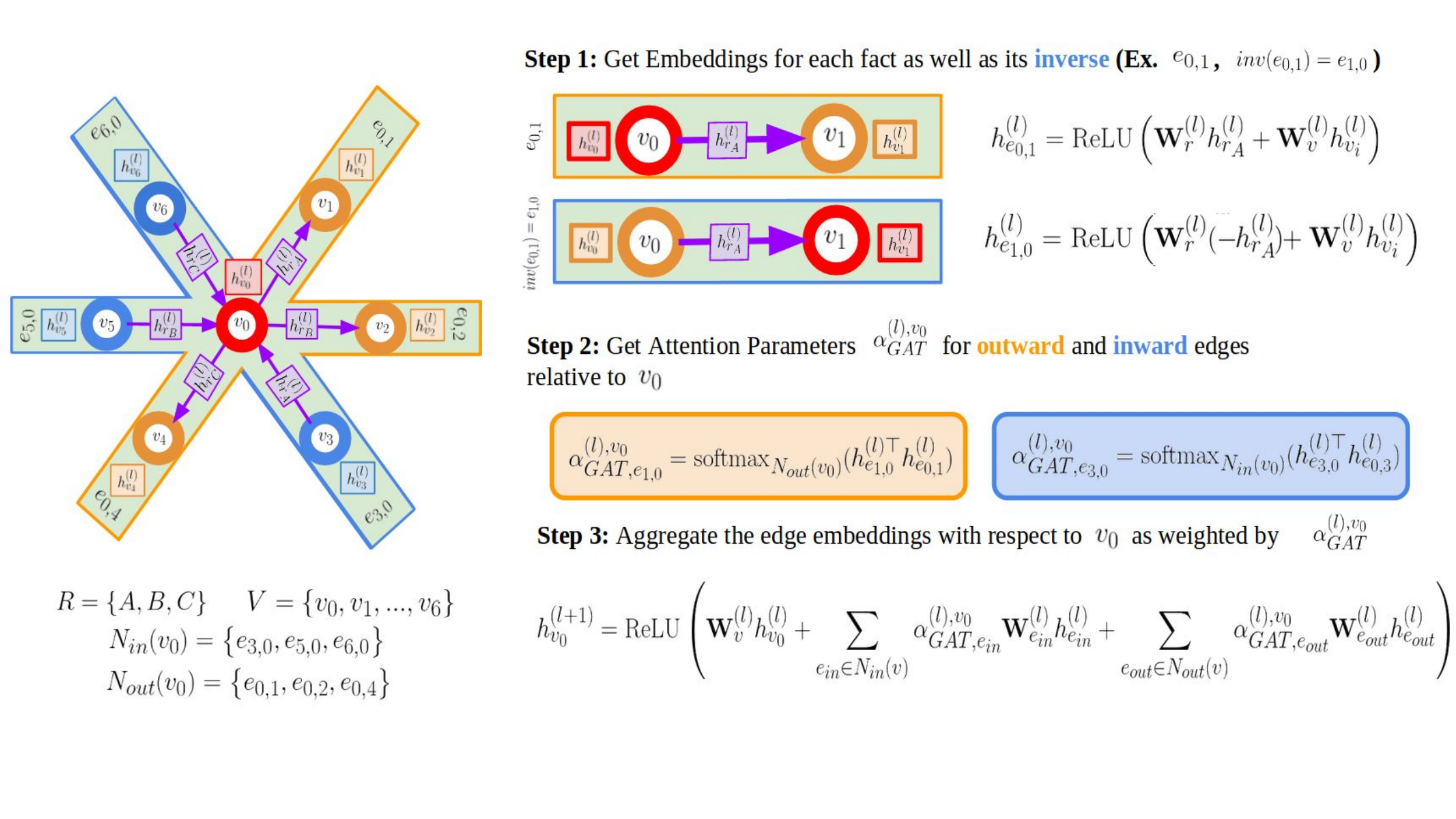}
  \caption{Message passing in proposed bi-directional graph attention mechanism.}
  \label{fig:GAT}
\end{figure}

The next step is aggregating the embeddings of the edges connecting to the node, i.e., $h^{l}_{e_{out}}$ and $h^{l}_{e_{in}}$. We apply two attention 
mechanisms to perform the aggregation and hence the model separately aggregates weighted sums of edge embeddings over each 
attention parameters: $\alpha^{(l)}_q$ and $\alpha^{(l)}_{GAT}$. The first attention parameter $\alpha^{(l)}_q$ is based on the normalized similarity between the relation 
embedding of an edge (i.e., $h_{r}^{(0)}$) and the question embedding (i.e., $h^{(0)}_q$) at layer 0: $\alpha^{(l)}_{q,e_{in}}=\text{softmax}(h_{r_{in}}^{0 \top} h_q^0)$ and
$\alpha^{(l)}_{q,e_{out}}=\text{softmax}(h_{r_{out}}^{0 \top} h_q^0)$. This captures the semantic similarity between the query and the inward and outward 
relations separately. The second attention parameter $\alpha^{(l)}_{GAT}$ is based on the similarity between a node and its neighbors with respect to the relationship 
between them. Assume $(v_i,r,v_j)$ is an edge between nodes $v_i$ and $v_j$ with relationship of type $r$.  The edge score is defined as the dot 
product between the edge embedding of the original direction and inverted direction $(v_j,r,v_i)$ and then normalized over all inward 
edges: $\alpha^{(l)}_{e_{in}}=\text{softmax}(h_{\overset{\shortrightarrow}{e}}^{(l)\top} h_{\overset{\shortleftarrow}{e}}^{(l)})$ and outward edges: $\alpha^{(l)}_{e_{out}}=\text{softmax}(h_{\overset{\shortrightarrow}{e}}^{(l)\top} h_{\overset{\shortleftarrow}{e}}^{(l)})$. Unlike 
graph attention \cite{ying_2018_neurips} this method addresses heterogeneous directed knowledge graphs. Finally, the model 
updates the node embedding (Figure \ref{fig:GAT}):
\begin{equation}
  h^{(l+1)}_v= \text{ReLU}\left(\textbf{W}_v^{(l)}h_{v}^{(l)} +
  \sum_{e_{in}\in{N_{in}(v)}}{\alpha^{(l)}_{e_{in}}\textbf{W}^{(l)}_{e_{in}}h^{(l)}_{e_{in}}}+
 \sum_{e_{in}\in{N_{out}(v)}}{\alpha^{(l)}_{e_{out}}\textbf{W}^{(l)}_{e_{out}}h^{(l)}_{e_{out}}}\right) 
  \tag{3}\label{eq:5}
\end{equation}
\subsection{Hierarchical Aggregation}
GNN models cannot learn hierarchical representations as they do not exploit the compositionality of graphs. Inspired by 
\cite{ying_2018_neurips} we define a linear layer to assign nodes into clusters and then use them to represent the nodes. 
This increases the receptive field by letting messages to pass among nodes belonging to the same cluster.  We compute the soft 
cluster assignments using a linear layer $\textbf{C}^{(l)}=\text{softmax}(\textbf{H}_{v}^{(l)}\textbf{W}_{c}^{(l)})$ where $H_v \in \mathbb{R}^{n_v \times d_v}$ is the node embedding matrix, 
$\textbf{W}_{c}^{(l)} \in \mathbb{R}^{d_v\times n_c}$ is a trainable parameter, and $\textbf{C}^{(l)}\in \mathbb{R}^{n_v\times n_c}$ is the normalized soft assignment matrix mapping $n_v$ nodes to $n_c$
clusters. We then compute the cluster centroids using $\textbf{H}_{c}^{(l)}=\textbf{C}^{(l)\top}\textbf{H}_{v}^{(l)}$and compute the cluster-based node representation using
$\textbf{H}_{vc}^{(l)}=\text{softmax}((\textbf{H}_{v}^{(l)}\textbf{W}_{c}^{(l)})^
{\top})\textbf{H}_{c}^{(l)}$. Finally, we concatenate the node representation in the final layer with all the cluster-based 
node representations from previous layers (i.e., similar to DenseNet \cite{huang_2017_cvpr}). We reduce the dimensionality using a trainable parameter  $\textbf{W}_{final} \in \mathbb{R}^{d_vL\times d_v}$ followed by a sigmoid function to produce the probability score per node.
\begin{equation}
\mathcal{Y}=\sigma\left(\textbf{W}_{final}\left[\left({\operatorname*{\parallel}_{l=0}^{L-1} \textbf{H}_{vc}^{(l)}}\right) \parallel \textbf{H}_{v}^{(L)}\right]\right)
\tag{4}
\end{equation}
\section{Results}
We implemented the model with Pytorch \cite{paszke_2017_pytorch} on a Nvidia DGX-1 server with 8 Volta GPUs and optimize it 
using Adam \cite{Kingma_2014_ICLR} with initial learning rate of 0.001 and batch size of 10 for 100 epochs with early stopping. The 
learning rate is decayed by a factor of 0.8 every 10 epochs. We also apply batch-normalization \cite{Ioffe_2015_icml} and dropout 
\cite{Srivastava_2014_jmlr}. We evaluated our model on the \textbf{WebQuestionsSP} dataset consisting of 4,737 natural 
language questions  (i.e., 3,098 training, 250 validation, and 1,639 test questions)  posed over Freebase entities \cite{liang_2017_acl}.
Following \cite{sun_2018_acl}, we apply the same pre-processing and report average F${_1}$ and Hits@1, as well as micro-average, and macro-average 
F${_1}$ scores. \cite{forman_2010_acm_sigkdd} suggests that micro-averaged F${_1}$ best represents the performance on imbalanced 
binary classification. 

Table 1 shows the performance of our model compared to other models that also feature early fusion of the knowledge graph and text. These include 
Key-Value Memory Networks (KVMN) \cite{das_2017_acl} and GRAFT-Net \cite{sun_2018_acl}. The results suggest that our model outperforms 
GRAFT-Net with an absolute increase in all metrics. To investigate the effect of the proposed methods we performed an ablation study by masking 
each introduced component and training and evaluating the model. The results in Table 2 (Appendix) shows the effect of each component and suggest 
that all introduced components contribute to the performance. 
     \begin{table}[t]
            \centering
    		\begin{tabular}{lcccc}
        		\toprule
        		\textbf{Model} & \textbf{F$^{micro}_1$} & \textbf{F$^{macro}_1$} & \textbf{F$^{avg}_1$} & \textbf{Hits@1} \\
        		\midrule
        		KVMN \cite{das_2017_acl}  &-&-& 30.9 & 40.5\\
        		GRAFT-Net \cite{sun_2018_acl}  & 66.0 & 61.0 & 60.4 & 67.8\\
        		\midrule
                 \textbf{Ours} & \textbf{72.6} & \textbf{62.5} & \textbf{61.7} & \textbf{68.2} \\
        		\bottomrule
   			\end{tabular}
   			\vspace{0.2cm}
   			\caption{Comparative results on the WebQuestionsSP benchmark.}
   			\vspace{-0.3cm}
    \end{table}

\section{Conclusion}
We introduced a relational GNN with bi-directional attention and hierarchical representation learning for 
open-domain question answering that jointly learns to represent the query, knowledge graph, and documents. The 
experiments showed that our model achieves state-of-the-art performance on WebQuestionsSP. For future directions, 
we are planning to expand our model towards cross-modal question answering benchmarks such as Fact-based Visual 
Question Answering (FVQA) \cite{wang2018fvqa}.

\small
\bibliographystyle{plain}
\bibliography{biblo}

\section {Appendix} 
     \begin{table}[!htb]
			\centering
    		\begin{tabular}{lcccc}
        		\toprule
        		\textbf{Masked Component} &  \textbf{F$^{micro}_1$} &
        		\textbf{F$^{macro}_1$} &
        		\textbf{F$^{avg}_1$} & \textbf{Hits@1} \\
       	 		\midrule
                Document Relations & 69.3 & 61.1 & 59.0 & 69.3 \\
        		Bi-Directional Attention  & 67.9 & 60.6 & 58.5   & 66.9 \\
        		Graph Coarsening & 67.9& 60.8 &58.7 & 67.8 \\
        		\midrule
                No Mask &\textbf{72.6} & \textbf{62.5} & \textbf{61.7} & \textbf{68.2} \\
        		\bottomrule
    		\end{tabular}
     		\vspace{0.2cm}
    		\caption{Effect of components on the performance of the proposed model.}
    \end{table}
\label{others}

\end{document}